 \documentclass[5p,times]{elsarticle}
\usepackage{nccmath}
\usepackage{lineno,hyperref}
\modulolinenumbers[5]
\usepackage[ruled,vlined]{algorithm2e}
\usepackage{setspace}
\usepackage{amssymb}
\usepackage{multirow}
\usepackage{booktabs} 
\usepackage{multicol}
\journal{Journal of \LaTeX\ Templates}

\begin{document}

\begin{frontmatter}


\title{Sparse-Dyn: Sparse Dynamic Graph Multi-representation Learning via Event-based Sparse Temporal Attention Network}





\author[mymainaddress]{Yan Pang}
\author[mymainaddress]{Chao Liu\corref{mycorrespondingauthor}}

\cortext[mycorrespondingauthor]{Corresponding author}
\ead{chao.liu@ucdenver.edu}

\address[mymainaddress]{Department of Electrical Engineering, University of Colorado Denver, CO}

\begin{abstract}
Dynamic graph neural networks have been widely used in modeling and representation learning of graph structure data. Current dynamic representation learning focuses on either discrete learning which results in temporal information loss or continuous learning that involves heavy computation. In this work, we proposed a novel dynamic graph neural network, Sparse-Dyn. It adaptively encodes temporal information into a sequence of patches with an equal amount of temporal-topological structure. Therefore, while avoiding the use of snapshots which causes information loss, it also achieves a finer time granularity, which is close to what continuous networks could provide. In addition, we also designed a lightweight module, Sparse Temporal Transformer, to compute node representations through both structural neighborhoods and temporal dynamics. Since the fully-connected attention conjunction is simplified, the computation cost is far lower than the current state-of-the-arts. Link prediction experiments are conducted on both continuous and discrete graph datasets. Through comparing with several state-of-the-art graph embedding baselines, the experimental results demonstrate that Sparse-Dyn has a faster inference speed while having competitive performance.

 
\end{abstract}


\end{frontmatter}

\nolinenumbers

\begin{figure*}[ht]
\vskip 0.2in
\begin{center}
\centerline{\includegraphics[width=\linewidth]{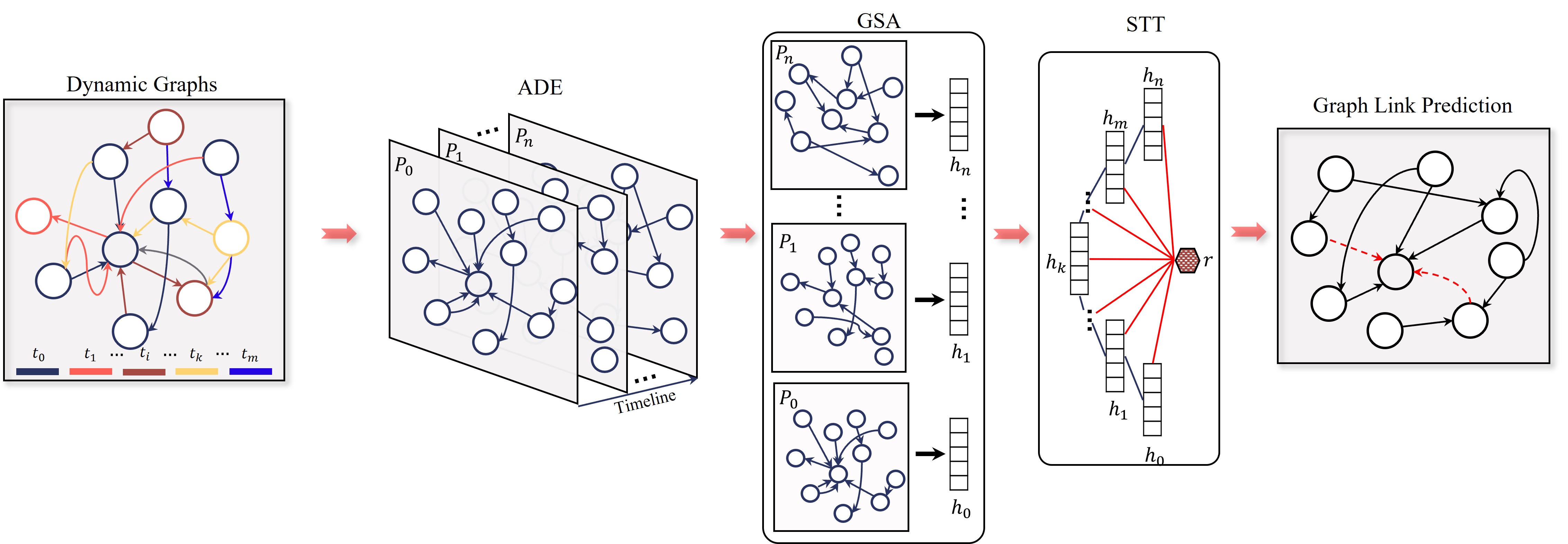}}
\caption{The overall architecture of Sparse-Dyn includes three main parts: Adaptive Data Encoding (ADE), Graph Structural Attention (GSA) and Sparse Temporal Transformer (STT).}
\label{fig:top}
\end{center}
\vskip -0.2in
\end{figure*}

\section{Introduction}
Dynamic graph neural networks (DGNNs) have seen a notable surge of interest with the encouraging technique for learning complicated systems of relations or interactions over time. Since DGNNs append an additional temporal dimension to accumulate the variation of embedding or representations, they are powerful tools to employ in diverse fields, such as social media \cite{perozzi2014deepwalk}, bio-informatics \cite{grover2016node2vec}, knowledge bases \cite{wang2014knowledge}, brain neuroscience \cite{goering2020fostering}, protein-protein interaction networks \cite{fout2017protein}, recommendation system \cite{ying2018graph}, etc.

In order to deal with the complicated time-varied graphs, it is necessary and crucial to preprocess the raw dynamic graph representations, which record all continuous evolution of the graph over time, such as node emerging/disappearing and link addition/deletion \cite{cui2018survey, cai2018comprehensive, kazemi2020representation, barros2021survey, skardinga2021foundations}. Current researches \cite{sankar2020dysat, xu2020inductive, ma2020streaming, zhou2018dynamic, goyal2020dyngraph2vec, chen2018gc, kumar2019predicting} refine the raw dynamic representations to two main branches, dynamic continuous and dynamic discrete graphs. The raw representations are projected to a single 2D temporal graph for the former graphs, storing the most information in graph evolution. However, the corresponding dynamic continuous networks are considerably complicated, which involves heavy computation \cite{skardinga2021foundations}. For discrete graphs, the structural representations are sampled to graph snapshots at regular time intervals, such as one day, over time. Although the developing networks are easier than the continuous ones, the temporal information is lost much more \cite{barros2021survey}. We hope to find an efficient way to encode the raw dynamic graph representations, which can alleviate the temporal information loss and simplify the evolved network in future representation learning. Thus, one of our primary contributions, Adaptive Data Encoding (ADE), is proposed to adequately project the temporal information into a sequence of event-based patches with equal amounts of temporal-topological structural patterns for avoiding information loss.

DGNNs extract and analyze patterns for graph learning along temporal dimension on the refined dynamic temporal graphs. It is crucial to have an efficient and powerful network under specific tasks in this step. Some researches \cite{ma2020streaming, chen2018gc, kumar2019predicting} utilize recurrent neural networks (RNNs) to scrutinize representations on the sequence of dynamic graphs. However, RNN-based DGNNs are more time-consuming and inadequately handle sequential time-dependent embedding with increasing moments of time-steps. Since the transformer-based approaches \cite{sankar2020dysat, xu2020inductive} adaptively designate divergent and interpretable attention to past embedding over time, the performance is better than RNN-based DGNNs on the long-time duration. However, because the standard transformer \cite{vaswani2017attention} contains fully-connected attention conjunction, which causes the heavy computation on a time-dependent sequence \cite{guo2019star}. We hope to simplify and effectively convey temporal information along the time dimension and achieve acceptable performance under inductive and transductive link prediction tasks. Thus, a lightweight module, Sparse Temporal Transformer (STT), is proposed to compute the temporal information with far lower costs by a simplified sparse attention conjunction under both graph tasks.

Figure \ref{fig:top} illustrates the overall architecture of the Sparse-Dyn, which contains three main components: ADE for adaptive encoding the raw dynamic continuous graph-based data, graph structural attention (GSA) for investigation on local structural patterns on patches, and STT for graph evolution capture of global temporal patterns over time.

In order to evaluate our proposed model, we conduct experiments on two continuous datasets under inductive link prediction tasks. The experiments demonstrate that Sparse-Dyn significantly outperforms state-of-the-art networks on the continuous graph datasets. In addition, we also designed an abbreviated version of Sparse-Dyn with only GSA and STT. This abbreviated version is then utilized to learn the representation on the discrete datasets under both inductive and transductive link prediction tasks. Further experiments show this version is still faster, more efficient, and have higher accuracy on four discrete dynamic graph datasets. 

The contributions in this paper are summarized as follows:
\begin{itemize}
\item The Sparse-Dyn is proposed to trade off the accuracy and efficiency on both dynamic continuous and discrete representations under link prediction tasks.
\item We recommend a new approach, Adaptive Data Encoding, to preprocess the raw dynamic graph representations. The ADE can alleviate the information loss in the process and simplify the evolved network in future representation learning.
\item We propose a lightweight temporal self-attentional module called Sparse Temporal Transformer. The STT-based Sparse-Dyn can substantially reduce the computation by comparing RNN-based and standard transformer-based solutions on both continuous dynamic graph datasets.
\item The abbreviated version of Sparse-Dyn, comprised only GSA and STT, can also be utilized on the discrete dynamic graph datasets. The experiments consistently demonstrate superior performance for Sparse-Dyn over state-of-the-art approaches under inductive and transductive link prediction tasks.
\end{itemize}


\section{Related Work} \label{RW}
\subsection{Graph Representations}
In general, graph representations can be categorized into two distinct levels: static and dynamic. The former includes only structural information, while the latter includes another critical parameter: time. The raw dynamic representation contains node interactions \cite{latapy2018stream}, and timestamped edges \cite{barros2021survey}, where instantaneous events are recorded into the raw graph representations, such as creation and removal of nodes and edges. 

Current researches mainly focus on two branches to prepossess the raw representations: dynamic continuous and discrete graphs \cite{cui2018survey, cai2018comprehensive, barros2021survey, skardinga2021foundations}. The dynamic continuous graphs store the most information by projecting the raw representations to a 2D temporal graph, which is also a specific static graph appended with temporal information \cite{kazemi2020representation}. However, the corresponding networks are complicated because they have to extract temporal information at each moment, which is the common issue of the dynamic continuous graphs \cite{skardinga2021foundations}.

For the dynamic discrete graphs, current research \cite{ cai2018comprehensive, kazemi2020representation} group graph embedding with a certain temporal granularity over time. The discrete graphs include discrete equal time intervals, which can be represented with multiple snapshots along temporal dimension \cite{taheri2019learning}. Because the temporal information is sampled at the discrete moments, such as one day/month, to several graph snapshots, the discrete representation is less complicated than a continuous representation \cite{cui2018survey}. However, this kind of graph tracking manner causes more information loss in the processing. Also, the distribution of events or interactions among different temporal windows is not homogeneous, which leads to an imbalance of temporal information among divergence graph snapshots.

In order to find an efficient temporal encoding approach to alleviating the temporal information loss and simplify the evolved network in future representation learning, we propose the event-based ADE module to adaptively encode the temporal information and determine the optimum number of temporal patches on the time dimension. Unlike traditional representations, our temporal patches contain an equal amount of temporal-topological structural patterns, which is fair and effective for future representation learning.

\begin{table*}[t]
\caption{General Notations}
\label{tab:notations}
\vskip 0.15in
\begin{center}
\begin{scriptsize}
\begin{sc}
\begin{tabular}{c|c|c|c}
\toprule
Notations & Description & Notations & Description \\
\midrule
DG & Dynamic Graphs & S & Snapshots \\
A & Adjacency Matrix & $X$ & Node Feature Vectors \\
E & Edges / Connections & N & Nodes / Objects\\
$e_{um} $ & Edge between Node u and m & $u$ & Center Node \\
$h$ & Structural Representation & $E$ & Embedding / Token Representation \\
$R$ & Frequency of Events & $W$ & Learnable Parameters \\
PE & Position Embedding & $p$ & Position-aware Structural Representation \\
$c$ & Context Information & $z$ & Time-dependent Structural Representation \\
$r$ & Relay Representation & L & Loss Function \\
\bottomrule
\end{tabular}
\end{sc}
\end{scriptsize}
\end{center}
\vskip -0.1in
\end{table*}

\subsection{Dynamic Graph Neural Network}
Since dynamic graph representations append a time dimension on the static ones, the RNN-based DGNNs \cite{goyal2020dyngraph2vec, hajiramezanali2019variational} are considered to summarize temporal information over time. However, the computation of RNN-based DGNNs is expensive because RNNs need a large amount of graph data for training. Moreover, it scales poorly on a long-time temporal dimension \cite{sankar2020dysat}. In order to solve this issue, the transformer-based DGNNs \cite{xu2020inductive, sankar2020dysat} are introduced to deal with the temporal information along the time dimension. TGAT \cite{xu2020inductive} introduces temporal constraints on neighborhood aggregation methods and utilizes a temporal graph attention layer to aggregate temporal-topological features on the continuous graph datasets. DySAT \cite{sankar2020dysat} generates a dynamic representation by joint self-attention of both structural and temporal information on discrete graph datasets. However, the common problem is that their computation is enormous on a long temporal sequence due to the fully-connected attention conjunction of the standard transformer. Under graph representation learning tasks, dynamic graph networks should achieve the desired trade-off between accuracy and efficiency. In order to achieve this target, our proposed Sparse-Dyn which contains a lightweight module, STT. Instead of the fully-connected attention conjunction, the information is only conveyed among 1-hop neighbors and the relay node. Experiments show that such a module significantly reduces the inference time and still achieves a better performance than the state-of-the-art approaches on both continuous and discrete representations.

\section{Preliminaries} \label{PR}
This section illustrates the terminology and preliminary knowledge. Table \ref{tab:notations} denotes various terminologies.

\textsc{Definition 1}. \textbf{Dynamic Graph Neural Network}. \textit{In general, a dynamic graph, $DG = (A, X; T)$, contains two main aspects: structural patterns $(A, X)$, and temporal information $T$.}

\textsc{Definition 2}. \textbf{Graph Link Prediction}. \textit{Given $DG$, the task is to estimate the link status among nodes by analyzing the aggregated information on both nodes at one moment.}

\textsc{Definition 3}. \textbf{Inductive Learning}. \textit{Given $DG$, DGNNs can only investigate the graph information from the beginning to the moment $T-1$. The analyzed representations are utilized to predict the future links at $T$. This task is prevalent and crucial since it makes predictions on unseen nodes and links in the future.}

\textsc{Definition 4}. \textbf{Transductive Learning}. \textit{Given $DG$, DGNNs can observe all nodes from the beginning to the end, $T$. The models learn representation and make the predictions at each snapshot or moment.}

\section{Methods} \label{ME}
The Sparse-Dyn consists of three components connected serially, ADE, GSA, and STT, as shown in Figure \ref{fig:top}. In order to make the distribution of the events uniformly along a temporal dimension and alleviate the disturbance from irrelevant messages to crucial ones in the future graph learning, ADE adequately encodes the temporal information to a sequence of patches with an equal amount of temporal-topological structural patterns by investigating the frequency of events. The GSA module extracts the local structural representations with a self-attention layer on each temporal patch. The learned time-dependent structural representations are sent to the lightweight module, STT, to capture the global graph evolution along the temporal dimension. 

\begin{figure}[ht]
\vskip 0.2in
\begin{center}
\centerline{\includegraphics[width=\linewidth]{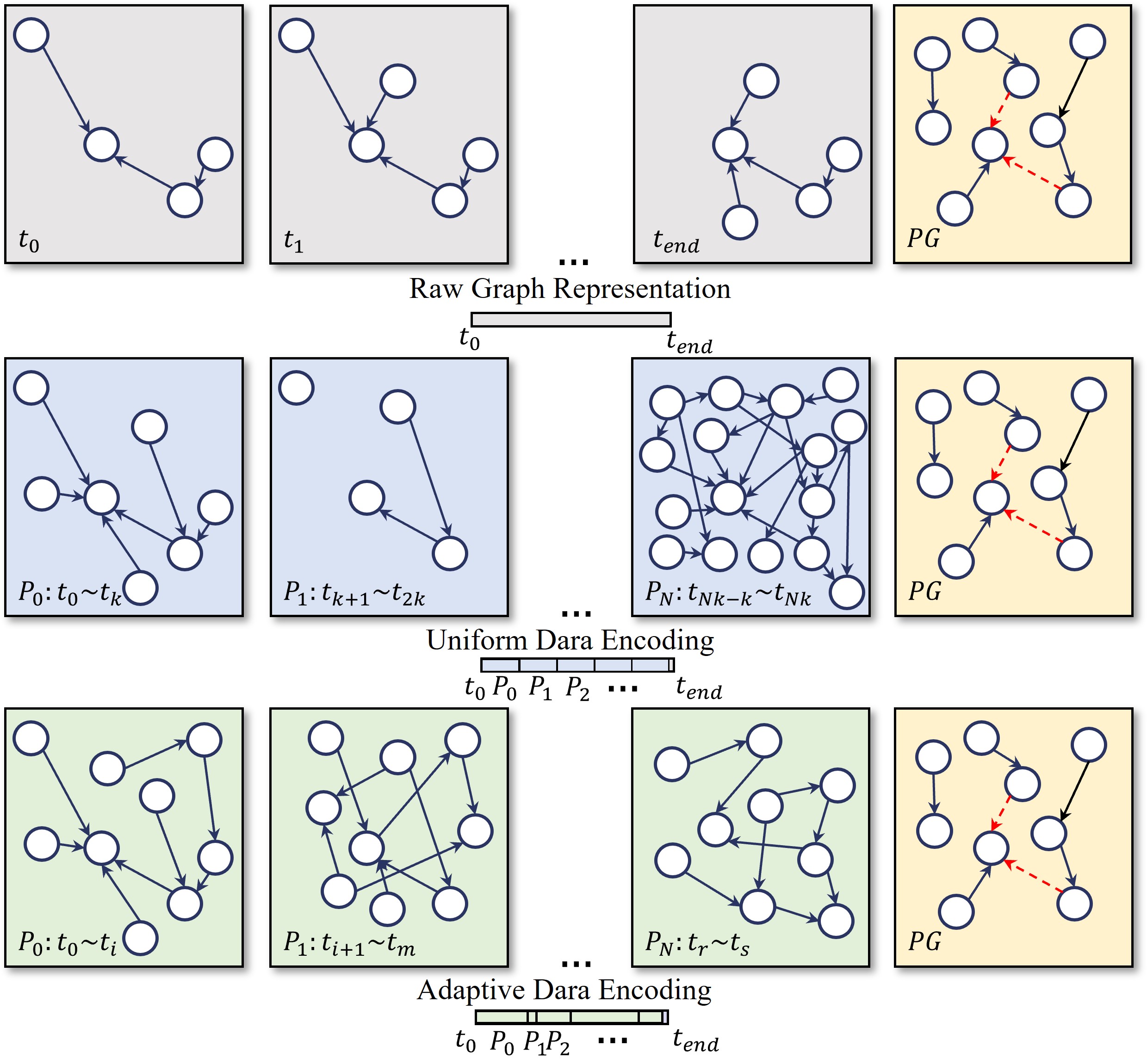}}
\caption{Different encoding approaches. Top row: raw graph representation; Second row: uniform data Encoding approach encodes the time-dependent representation by the same time interval uniformly; Bottom row: Adaptive data Encoding approach adaptively encodes the whole raw representation to a sequence of event-based patches. Because each patch contains an equal amount of temporal-topological patterns, the processing is high-efficient in parallel computing. PG indicates the predicted graph.}
\label{fig:ate}
\end{center}
\vskip -0.2in
\end{figure}

\subsection{Adaptive Data Encoding}\label{ADE}
Since dynamic continuous graph stores almost events over time, it can be regarded as a particular static graph with an additional temporal dimension. Current researches \cite{ma2020streaming, trivedi2019dyrep,han2019graph} project the topological graph structures and node features from the raw representation to a 2D temporal continuous graph for future representation learning. However, the data processing is low-efficiency because they pay much attention to the inoperative information for the target node \cite{kazemi2020representation}. The continuous representation of these networks is far complicated \cite{skardinga2021foundations}. In contrast, the dynamic discrete graph simplifies the data processing by sampling the structural representations to graph snapshots at regular time intervals. The developing discrete networks are less complicated than the continuous ones \cite{skardinga2021foundations}. However, the temporal information is lost much more.

\begin{figure}[ht]
\vskip 0.2in
\begin{center}
\centerline{\includegraphics[width=\linewidth]{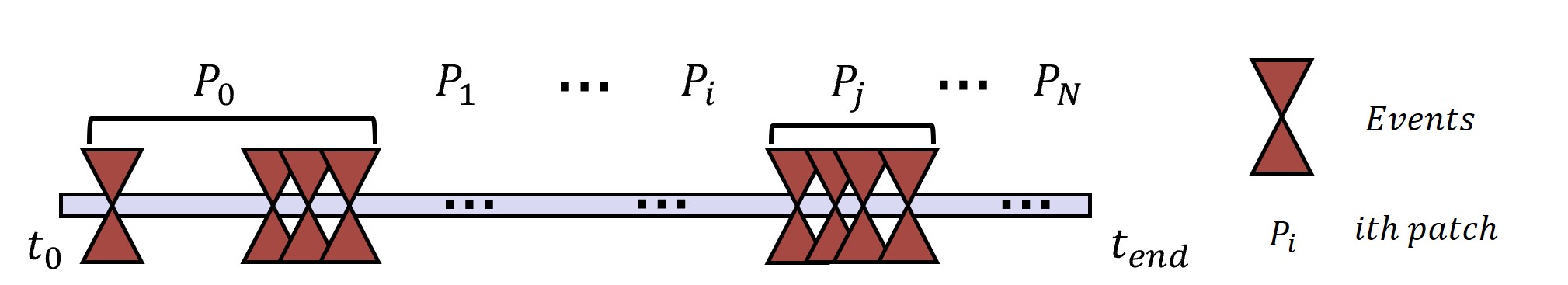}}
\caption{The ADE adaptively encodes temporal information into a sequence of patches by events along the temporal dimension. Each temporal patch contains an equal amount of temporal-topological patterns with the others.}
\label{fig:patch}
\end{center}
\vskip -0.2in
\end{figure}

We hope to find an efficient way to encode raw dynamic representations and reduce the information loss in this processing. Thus, an ADE is proposed to adaptively encode temporal information into a set of event-based temporal patches with an equal amount of temporal-topological structure. While avoiding the use of snapshots which causes information loss, it also achieves a finer time granularity close to what a dynamic continuous graph could provide.

Figure \ref{fig:ate} exposes the comparison of three encoding approaches: raw graph representation without encoding, uniform data encoding (UDE), and ADE. Based on the raw time-dependent representations with superabundant details, the UDE separates the graph patterns into several patches by the same temporal intervals, such as one day, et al. However, the distribution of temporal-topological patterns is not homogeneous of the encoded patches because the events have happened irregularly along the time dimension. The computation on some patches is expensive because it takes more to deal with the complicated patch structure; meanwhile, the calculation is light on those with fewer events. Thus, it is inefficient to process the patches with different amounts of structural patterns in parallel computing.

Instead of separating by a regular interval, the raw data are divided into $N$ event-based patches along a temporal dimension by ADE as shown in Figure \ref{fig:patch}. The amount of structural patterns or embedding of patch $P_{i}$ is equal to the one of patch $P_{j}$. In this procedure, it is crucial to balance the number of patches, $N$, and the quality of structural embedding of each patch along the time dimension. Thus, a cost function of ADE is utilized to adaptively determine $N$ and the embedding of patches in Equation \ref{la1} and \ref{la2}.

\begin{equation}\label{la1}
  min (L_A)=\phi (E, \sum_{n=1}^{N}E_{n})+\tau log\frac{R}{\Delta }
\end{equation}

\begin{equation}\label{la2}
  \phi\left ( E_i - E_j\right ) - \epsilon =0
\end{equation}
Where $E$ is the embedding of the raw graph representations, and $\tau$ is a constant parameter. The $R$ is the number of total events, and $N = \frac{R}{\Delta}$ is the number of patches. The $\phi$ is the Pearson Correlation Coefficient \cite{benesty2009pearson} to measure the linear correlation of embedding between every two patches. The $\epsilon$ is a tiny constant parameter to ensure that each patch contains an equal amount of temporal-topological patterns.

In Equation \ref{la1}, the first item is to minimize the difference between the quality of final projected representations of all patches. Ideally, $N = 1$ indicates the difference is eliminated, like the continuous representations. The second item is encouraged to increase a large value of $N$. By minimizing the cost function of $L_A$, an optimum balance between $N$ and $E_i$ can finally be obtained.

\begin{figure*}[ht]
\vskip 0.2in
\begin{center}
\centerline{\includegraphics[width=\linewidth]{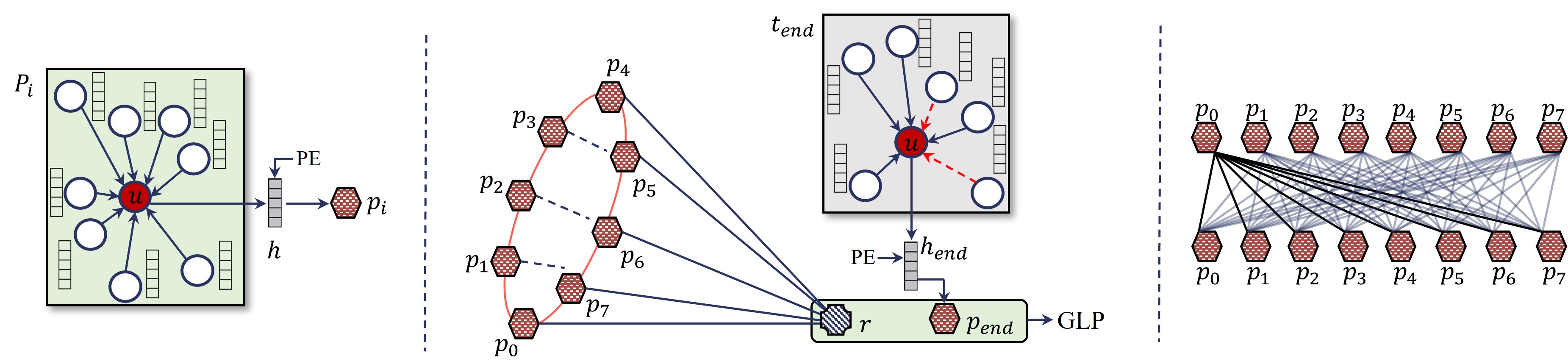}}
\caption{Left: GSA: The local structural patterns are extracted and added with the position embedding on each time-dependent patch in the GSA module. Middle: the lightweight STT: The global temporal patterns are investigated over time. Each patch is only connected to its two adjacent patches and the relay one. The updated relay patch is utilized to predict the predicted patch's link. Right: Patch conveys information to each other due to the fully-connected attention conjunction of the standard transformer, which causes heavy computation.}
\label{fig:arch}
\end{center}
\vskip -0.2in
\end{figure*}

\subsection{Graph Structural Attention}\label{GSA}
Once the raw graph representations are encoded to the optimal $N$ time-dependent patches, the next step is to extract the local structural patterns on each encoded patch. The input of GSA is a set of node representations of the current patch. Inspired by GAT \cite{velivckovic2017graph}, a node self-attention matrix, $\alpha_{um}=softmax(e_{um})$, is learned to determine the relevance between neighbors and the center node, $u$, on the time-dependent patch.

\begin{equation}\label{eum}
  e_{um}=\sigma (A_{um}\cdot r[W_{p}X_{u};W_{p}X_{m}])
\end{equation}

Where $e_{um}$ indicates the relevance of neighbor node $m$ to the center node $u$. The $\sigma$ is the exponential linear unit (ELU) activation function \cite{trottier2017parametric}. The $A$ is the adjacency matrix of the patch to indicate the linking relation of the current patch, The $\gamma$ indicates the self-attention mechanism, and $W_p$ is the weight matrix of the patch $p$. The $x_{u}$ and $x_{m}$ are the node representation of the center node $u$ and neighbor node $m$. Then, the activation function $\psi$ is applied to get the non-linear node representation of the current patch in Equation \ref{hu}.

\begin{equation}\label{hu}
  h_{u}=\psi \left ( \sum_{m\sqsubseteq N_{u}}\alpha_{um}\cdot W_{p}X_{m}\right )
\end{equation}

Where $\psi$ is the Gaussian Error Linear Unit (GELU) \cite{hendrycks2016gaussian} for the final output representations, $h_u \in  \mathbf{R}^d$ is the patch embedding, and $d$ is the updated feature dimension.

The final step of GSA is to add the position embedding of time-dependent patches, which embed the absolute temporal position of each patch, as shown in Figure \ref{fig:arch} Left. Thus, the output of GSA, $p_i$, contains both local structural patterns and temporal position information of the current patch.

\subsection{Sparse Temporal Transformer }\label{STT}
The objective of STT is to gather the global evolution of structural patterns on each patch from GSA along the time dimension. Current transformer-based DGNNs \cite{xu2020inductive, sankar2020dysat} utilize the standard transformer to extract the temporal patterns and receive good accuracy on both continuous and discrete dynamic representations. However, due to the fully-connected attention conjunction as shown in Figure \ref{fig:arch} Right, the computation is expensive of these transformer-based DGNNs on a long time sequence. In order to reduce the computation, we design a lightweight module, STT, instead of the standard transformer to deal with the global temporal patterns of time-dependent patches.

Figure \ref{fig:arch} Middle illustrates the architecture of the STT module, which consists of $N$ time-dependent patches from GSA and a single relay patch. Each temporal patch is connected with two adjacent patches and the relay one on the STT. The functionality of the relay patch is to congregate and distribute the representation among all time-dependent patches. Thus, the STT module can learn global representations with the relay patch. By comparing the fully-connected attention conjunction, the advantage of our connection is that the computation to extract the temporal information is cut down by reducing the interaction times of patches.

The input of STT is a sequence of representations for a center node $u$ at all temporal patches. The embedding of the relay patch is initialized as the average of all time-dependent patches at the beginning. The context of $c_{i}(t)$ of the patch $i$ is updated by aggregating the representations from its two neighbor patch $i-1$ and $i+1$, the relay $r(t-1)$, the state of itself at last moment $z_{i}(t-1)$, and the embedding $p_{i}$.

\begin{equation}\label{cit}
  c_{i}(t)=\left [ z_{i-1}(t-1); z_{i}(t-1); z_{i+1}(t-1); p_{i}; r(t-1) \right ]
\end{equation}

The temporal self-attention function of the current state $z_{i}(t)$ of patch $i$ is defined as in Equation \ref{zit}.

\begin{equation}\label{zit}
  z_{i}(t)=\phi_{1}\left ( \beta _{i}(t) \cdot c_{i}(t)W_{i} \right )
\end{equation}

\begin{equation}\label{bit}
  \beta _{i}(t)=softmax\left ( \frac{z_{i}(t-1)W_{q} \cdot (c_{i}(t)W_{k})^{T}}{\sqrt{d}} \right )
\end{equation}

Where $\beta _{i}(t)$ is the self-attention coefficients of temporal patches, $W_{q}, W_{k}, W_{i}$ are learnable parameters, and $d$ is the feature dimension of $z_i$. A layer normalization operation \cite{krueger2016zoneout} is added after the transaction among all patches. 

Meanwhile, current state of relay patch $r(t)$ gather all representations from temporal patches $Z(t)$, and the state of itself at last moment $r(t-1)$ in Equation \ref{rt}.

\begin{equation}\label{rt}
  r(t)=\phi_{2}\left ( \lambda (t) \cdot \left [ r(t-1);Z(t) \right ] W'_{v}\right )
\end{equation}

\begin{equation}\label{lamdat}
  \lambda (t)=softmax\left ( \frac{r(t-1)W'_{q} \cdot (\left [ r(t-1);Z(t) \right ]W'_{k})^{T} }{\sqrt{d'}} \right )
\end{equation}

Where $\lambda (t)$ is the self-attention coefficients of relay patch, $W'_{q}, W'_{k}, W'_{i}$ are learnable parameters, and $d'$ is the feature dimension of the relay. Both $\phi_{1}$ and $\phi_{2}$ are non-linear activation function. Similarly, the layer normalization operation is also added after the transaction on relay patches. 

\subsection{Graph Link Prediction}
Graph link prediction is one of the core graph tasks, whose purpose is to forecast the connection among nodes based on node representations. In the inductive task, the representations of $T-1$ temporal patches are analyzed in the training processing. The network makes the link prediction to the unseen nodes on the final predicted graph (PG). In this procedure, we utilize the deep walk \cite{perozzi2014deepwalk} approach to sample some positive (connected links) and negative (unrelated links) on PG. A binary cross-entropy loss is to embolden positive cases to have similar representations while suppressing the negative ones in Equation \ref{loss_i}.

\begin{equation}\label{loss_i}
\begin{split}
  L=\sum_{u\in V} ( &\sum_{v\in N_{walk}(u)}-log\left ( \varphi (<e_{v},e_{u}>) \right ) \\ 
  -\omega _{n} \cdot &\sum_{v'\in P_{walk}(u)} log\left ( \varphi(1-<e_{v'},e_{u}>) \right ) ) 
\end{split}
\end{equation}

Where $\varphi$ is the non-linear activation function, $<\cdot>$ is the inner-production, $\omega _{n}$ is a constant fine-tuned hyper-parameter. The $ N_{walk}(u)$ is the sampled positive cases in fixed-length random deep walks on PG, while the $P_{walk}(u)$ is the sampled negative cases. 

In terms of the transductive task, the positive and negative cases are sampled at each temporal patch. Thus, the final loss function is to calculate the sum of all costs on each patch in Equation \ref{loss_t}.

\begin{equation}\label{loss_t}
\begin{split}
  L=\sum^{T_N}_{t=1} \sum_{u\in V} ( &\sum_{v\in N^{t}_{walk}(u)}-log\left ( \varphi (<e_{v}^{t},e_{u}^{t}>) \right ) \\ 
  -\omega _{n} \cdot &\sum_{v'\in P^{t}_{walk}(u)} log\left ( \varphi(1-<e_{v'}^{t},e_{u}^{t}>) \right ) ) 
\end{split}
\end{equation}

\section{Experiments} \label{EX}
\subsection{Datasets}
We experimentally validate Sparse-Dyn on six real-world dynamic graph datasets: two dynamic continuous and four dynamic discrete datasets. Table \ref{tab:dynamic} summarizes the statistics of the details of these six datasets.

\begin{table}[t]
\caption{Statistics of the dynamic graph datasets}
\label{tab:dynamic}
\vskip 0.15in
\begin{center}
\begin{scriptsize}
\begin{sc}
\begin{tabular}{c|ccc}
    \toprule
    Continuous & Nodes & Links & Time Duration(s)\\
    \midrule
    Reddit & 10984 & 672447 & 2678390 \\
    Wikipedia & 9227 & 157474 & 2678373 \\
    \midrule
    \midrule
    Discrete & Nodes & Links & Time Steps\\
    \midrule
    Enron & 143 & 2347 & 16 \\
    UCI   & 1809 & 16822 & 13 \\
    Yelp & 6509 & 95361 & 12 \\
    ML-10M & 20537 & 43760 & 13 \\
  \bottomrule
\end{tabular}
\end{sc}
\end{scriptsize}
\end{center}
\vskip -0.1in
\end{table}

\textbf{Reddit and Wikipedia} These two dynamic continuous graph datasets \cite{kumar2019predicting} describe the active users and their editions on Reddit and Wikipedia in one month. The dynamic labels represent the state of the user on their editions. Reddit contains 10984 nodes and 672447 links, while Wikipedia contains 9227 nodes and 157474 links.

\textbf{Enron and UCI} These two dynamic discrete graph datasets describe the network communications. Enron includes 143 nodes (employees) and 2347 links (email interactions), while UCI includes 1809 nodes (users) and 16822 links (messages).\cite{klimt2004enron, panzarasa2009patterns}

\textbf{Yelp and ML-10M} These two dynamic discrete graph datasets describe the bipartite networks from Yelp and MovieLens \cite{harper2015movielens}. The Yelp has 6509 nodes (users and businesses) and 95361 links (relationship), while ML-10M has 20537 nodes (users with the tags) and 43760 links (interactions).

\subsection{Inductive learning on continuous datasets}
These experiments compare different approaches on two continuous datasets under the inductive link prediction task. In these experiments, we compare Sparse-Dyn networks with another four approaches as the baselines: GAT-T \cite{velivckovic2017graph}, GraphSAGE-LSTM \cite{hamilton2017inductive}, Const-TGAT \cite{xu2020inductive}, and TGAT. When gathering the temporal information, GAT-T concatenates the time encoding to the graph structural features. GraphSAGE-LSTM considers Long Short-Term Memory (LSTM) to aggregate the temporal information over time. TGAT utilizes a temporal attention coefficient matrix to aggregate temporal representations. Const-TGAT pays the same temporal attention to collecting the temporal patterns. 

\begin{table*}
  \caption{Link prediction on dynamic continuous datasets. Left: Inductive learning task results (accuracy\%); Right: Inference time. The Sparse-Dyn* combines the functional time encoding component from TGAT and our STT component.}
  \label{tab:con_ilacc}
  \vskip 0.15in
  \begin{center}
  \begin{scriptsize}
  \begin{sc}
  \begin{tabular}{c|cccccc||ccc}
    \toprule
    Datasets & GAT-T & GraphSAGE-L & Const-TGAT & TGAT & Sparse-Dyn* & Sparse-Dyn & TGAT & Sparse-Dyn* & Sparse-Dyn\\
    \midrule
    Reddit & 90.24 & 89.43 & 88.28 & 90.68 & 92.83 & 88.65 & 30.164s & 23.187s & 18.956s \\
    Wikipedia & 84.76 & 82.43 & 83.60 & 85.28 & 87.36 & 83.55 & 15.377s & 11.637s & 9.623s \\
  \bottomrule
\end{tabular}
\end{sc}
\end{scriptsize}
\end{center}
\vskip -0.1in
\end{table*}

\begin{table*}
  \caption{Link prediction on dynamic discrete graph datasets. Top: Inductive learning task results (accuracy\%); Bottom: Transductive learning task results (accuracy\%). The Sparse-Dyn\S \  only consists of GSA and STT modules.}
  \label{tab:dis_ilacc}
  \vskip 0.15in
  \begin{center}
  \begin{scriptsize}
  \begin{sc}
  \begin{tabular}{c|cccccccc}
    \toprule
    Inductive & node2vec & GraphSAGE & GAT & DynamicTriad & DynGEM & DynAERNN & DySAT & Sparse-Dyn\S\\
    \midrule
    Enron    &  75.86  &  74.67 & 69.25 &  68.77 & 62.85 & 59.63 & 78.52 & \textbf{81.36}\\
    UCI      &  74.76  &  79.41 & 73.78 &  71.67 & 79.82 & 81.91 & 83.72 & \textbf{85.47}\\
    Yelp     &  65.17  &  58.81 & 65.91 &  62.83 & 66.84 & \textbf{73.46} & 69.23 & 72.59\\
    ML-10M   &  84.89  &  89.14 & 84.51 &  84.32 & 83.51 & 88.19 & 92.54 & \textbf{94.28}\\
  \bottomrule
  \toprule
    Transductive & node2vec & GraphSAGE & GAT & DynamicTriad & DynGEM & DynAERNN & DySAT & Sparse-Dyn\S\\
    \midrule
    Enron    &  83.05  &  81.88 & 75.97 &  78.98 & 69.72 & 72.01 & 86.60 & \textbf{87.94}\\
    UCI      &  80.49  &  82.89 & 81.86 &  80.28 & 79.82 & 83.52  & 85.81 & \textbf{87.53}\\
    Yelp     &  65.34  &  58.56 & 65.37 &  62.69 & 65.94 & 68.91 & 69.87 & \textbf{72.01}\\
    ML-10M   &  87.52  &  89.92 & 86.75 &  88.43 & 85.96 & 89.47 & 96.38 & \textbf{97.52}\\
  \bottomrule
\end{tabular}
\end{sc}
\end{scriptsize}
\end{center}
\vskip -0.1in
\end{table*}

\begin{table}
  \caption{Inference time (ms) on discrete graph datasets in the inductive learning task }
  \label{tab:dis_iltime}
  \vskip 0.15in
  \begin{center}
  \begin{scriptsize}
  \begin{sc}
  \begin{tabular}{c|cc}
    \toprule
    Datasets & DySAT & Sparse-Dyn\S \\
    \midrule
    Enron & 2.961 & 0.997 \\
    UCI   & 13.953 & 4.965 \\
    Yelp & 1360.31 & 509.62 \\
    ML-10M & 10678.36 & 3746.55 \\
  \bottomrule
\end{tabular}
\end{sc}
\end{scriptsize}
\end{center}
\vskip -0.1in
\end{table} 

Table \ref{tab:con_ilacc} shows the accuracy of these approaches under the link prediction task on two dynamic continuous datasets. It can be observed that the performances of transformer-based networks are better than RNN-based ones. With the temporal attention, the accuracy of TGAT can exceed 2.4\% and 1.68\% than the ones of Const-TGAT on Reddit and Wikipedia. 

TGAT consists of two main components, functional time encoding and a standard transformer. We modify the architecture of TGAT with STT instead of the standard transformer (ST) and name it as Sparse-Dyn*. In order to check the performance of ADE and STT separately on dynamic continuous representations, the performance of TGAT, Sparse-Dyn*(functional time encoding + STT), and Sparse-Dyn (ADE + STT) are analyzed in Table \ref{tab:con_ilacc}. Compared with TGAT and Sparse-Dyn*, the latter's accuracy achieves 92.83\% on Reddit and 87.46\% on Wikipedia, which surpasses 2.15\% and 2.04\% than the former. Meanwhile, the inference time of Sparse-Dyn* is less than TGAT, which demonstrates that STT is more effective by comparing the fully-connected connection of the standard transformer. Due to ADE, the inference time of Sparse-Dyn is further reduced by comparing with the time of Sparse-Dyn*. The Sparse-Dyn's accuracy is less than TGAT by 2.03\% and 1.73\% on Reddit and Wikipedia datasets because the functional time encoding component utilizes more details with temporal constraints of graph representations. However, our inference speed is only around 0.6 times that of TGAT on both continuous datasets, which is more competitive in practical usage. These experiments demonstrate the contribution of Sparse-Dyn consisting of both EDA and STT on dynamic continuous representations under the link prediction task.

\subsection{Inductive learning on discrete datasets}
The previous experiments demonstrate the power of Sparse-Dyn on dynamic continuous datasets. The Sparse-Dyn can also be utilized on discrete graph datasets. Since the discrete representations have several graph snapshots along temporal dimension, we compare our Sparse-Dyn\S, which only consists of GSA and STT, with another seven baselines: node2vec \cite{grover2016node2vec}, GraphSAGE, GAT, Dynamic Triad \cite{zhou2018dynamic}, DynGEM \cite{goyal2018dyngem}, DynAERNN \cite{goyal2020dyngraph2vec} and DySAT. The node2vec handles the second-order random walk sampling to grasp node representations. Dynamic Triad combines triadic closure to preserve both structural information and evolution patterns. DynGEM utilizes a deep autoencoder to generate non-linear embeddings of snapshots. DynGEM constructs both dense and recurrent layers to investigate the temporal graph evolution. DySAT extract node representations via fully-connected self-attention on both graph structural and temporal patterns.

Table \ref{tab:dis_ilacc} summarizes the results of these eight approaches on four dynamic discrete graph datasets. We find that the accuracy of DySAT exceeds the other state-of-the-art approaches, except Sparse-Dyn\S, under the link prediction task, which benefits from the fully-connected attention conjunction architecture of transform by extracting temporal patterns over time. This phenomenon demonstrates that the transformer-based DGNN outperforms the traditional graph learning approaches, including RNN-based DGNNs. By comparing DySAT and Sparse-Dyn\S, we found the accuracy of Sparse-Dyn\S is 81.36\%, 85.47\%, 72.59\%, and 94.28\% on Enron, UCI, Yelp, and ML-10M datasets, which are better than DySAT. Meanwhile, the inference time of Sparse-Dyn\S \ is much less than the time of DySAT on all four dynamic discrete datasets as shown in Table \ref{tab:dis_iltime}, which demonstrates that STT is also competitive and effective on dynamic discrete graph representations.

\subsection{Transductive learning on discrete datasets}
Besides previous experiments, we also evaluate our proposed network on dynamic discrete datasets under the transductive link prediction task. From Table \ref{tab:dis_ilacc} bottom, we observe the transformer-based DGNNs (DySAT and Sparse-Dyn\S) have better performances than RNN-based ones on all four discrete datasets, which also prove the self-attention architecture is powerful for transductive graph learning. As a result, the accuracy of Sparse-Dyn\S achieves 87.94\%, 87.53\%, 72.01\%, and 87.52\% on Enron, UCI, Yelp, and ML-10M separately, which also demonstrates the improvements delivered by our innovative architecture of Sparse-Dyn.

\section{Conclusion} \label{CON}
This paper proposes a sparse dynamic graph neural network, Sparse-Dyn, that trade-offs the accuracy and efficiency under inductive and transductive link prediction tasks. Sparse-Dyn consists of three main components: ADE, GSA, and STT. The ADE module adaptively encodes temporal information into a sequence of patches with an equal amount of temporal-topological structure, which reduces the information loss in the projection processing due to adaptive generation with a more delicate time granularity. Also, it simplifies the evolved network in future representation learning. The GSA module learns the local structural representations on each encoded patch along the temporal dimension. The lightweight STT is utilized to extract global temporal patterns over time. Benefiting from the information delivery on the simplified architecture, the STT-based Sparse-Dyn can substantially reduce the computation by comparing RNN-based and standard transformer-based solutions on both continuous dynamic graph datasets. The Sparse-Dyn is evaluated on two dynamic continuous and four dynamic discrete graph datasets. The results illustrate that Sparse-Dyn is competitive and efficient in inference speed and performance.


\section{Appendix}

\subsection{Event-based Data Encoding}\label{de}
\begin{figure}[h]
  \centering
  \includegraphics[width=\linewidth]{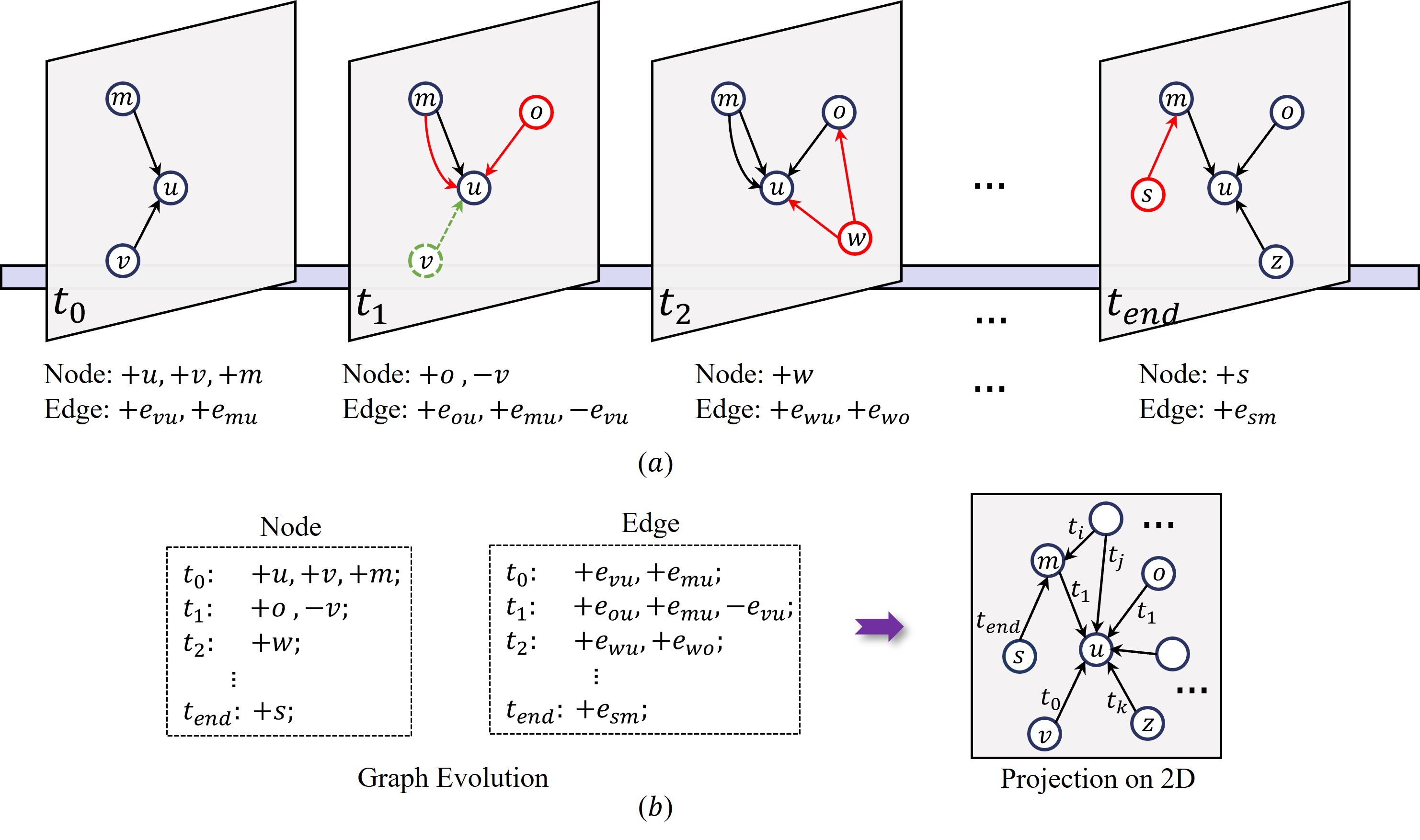}
  \caption{Visual illustration for projection from the raw dynamic continuous representations. a). The generation process of a continuous temporal graph and its snapshots at each moment. The solid red line represents an addition, and the dash green one represents deletion. b). The temporal information and the final state of the projected temporal graph. Some temporal information for nodes and multi-edges is lost.}
  \label{fig:raw}
\end{figure}

This section is to discusses the event-based data encoding approaches on raw dynamic representations. As shown in Figure \ref{fig:raw}.a, the raw dynamic continuous representation is a sequence of particular static graphs along time dimension, which stores all events, such as node emerging, node disappearing, link addition, link removing, et al. The crucial information is the recorded time-dependent events in the raw representations. Before the representation learning, the raw representations should be prepossessed. For the continuous graphs, the general approach is to project the raw representations to a single 2D temporal graph, as shown in Figure \ref{fig:raw}.b. The primary issue is that some temporal information is lost on the single graph, including node or edge vanishing and multi-edge situations. Also, the developing networks are complicated and contain heavy computation because they have to extract the temporal information at each moment on the dynamic continuous graph.

In order to alleviate the above issues, a straight thought is to convert the raw continuous representations to several small temporal graphs with temporal intervals instead of the large single one. Unlike dynamic discrete graphs that sample the representations at each discrete interval, we project all events in each period to temporal patches. Each encoded graph patch holds the temporal information in the duration with the same time interval, such as one day, one week, one month, et al. However, it is impossible to guarantee that events are uniformly distributed along the time dimension. With the uniform data encoding approach, some patches contain superabundant details due to more events in the corresponding duration and vice versa. 

\begin{figure}[h]
  \centering
  \includegraphics[width=\linewidth]{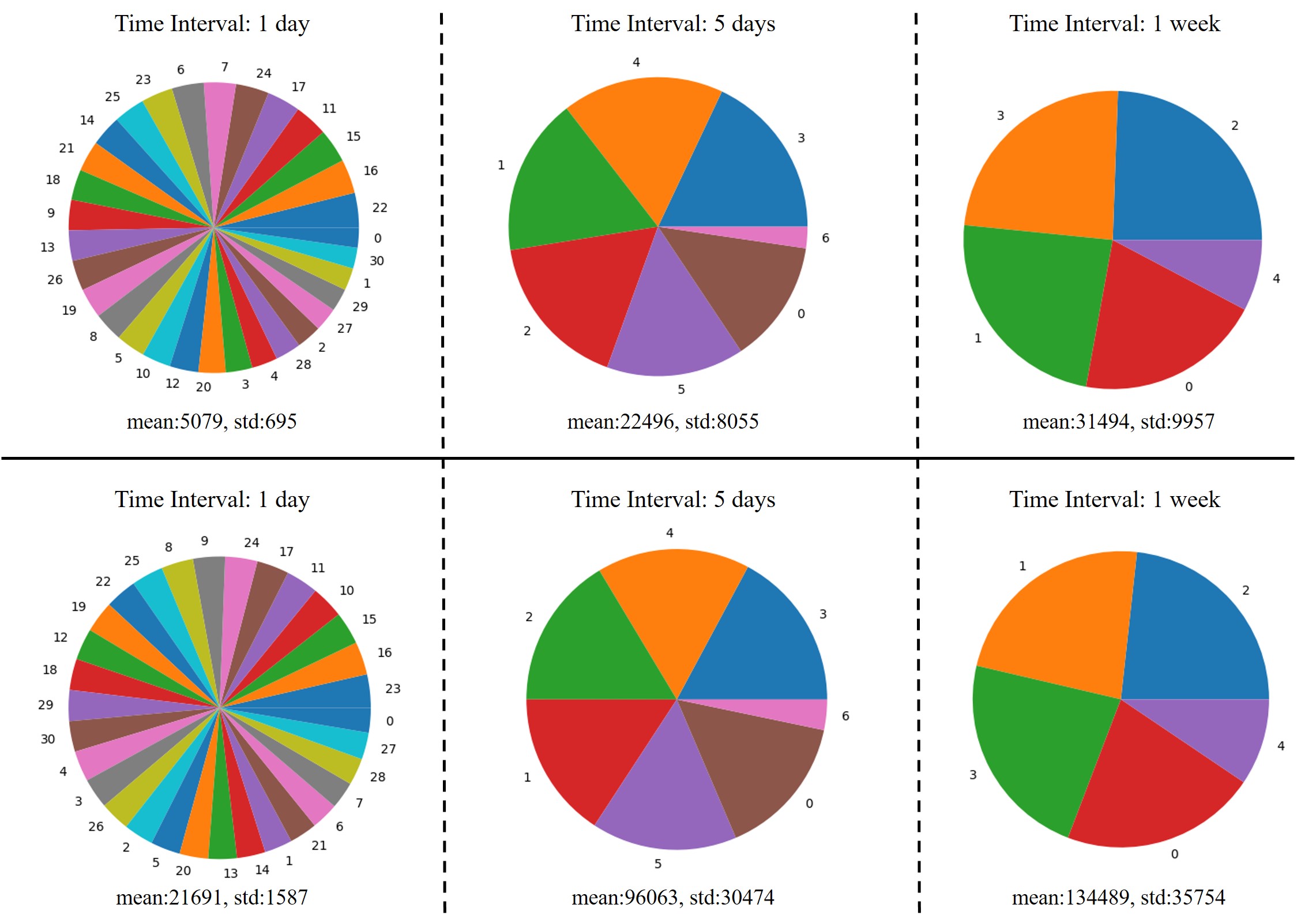}
  \caption{The events distribution is inhomogeneous with different uniform time intervals on Wikipedia and Reddit continuous datasets. Top: Wikipedia; Bottom: Reddit.}
  \label{fig:interval}
\end{figure}

As shown in Figure \ref{fig:interval}, We design some experiments to verify the above phenomenon. The top row of Figure \ref{fig:interval} is the distribution of events at each temporal patch on Wikipedia with three uniform time intervals: one day, five days, and one week. It can be observed that the distribution of events is not homogeneous at all three patches. It is becoming increasingly apparent with a longer time interval. The standard variance of the patch with a one-week interval is 9957, far outweighing the one with a one-day interval. A similar phenomenon emerges on Reddit, as shown in the bottom row of Figure \ref{fig:interval}. The calculation is also inefficient on these inhomogeneous patches in the future representation learning in parallel computing.

Our proposed adaptive data encoding approach adaptively encodes temporal information into a sequence of patches by events. The advantage of this approach is to avoid using snapshots to cause information loss and achieve a finer time granularity, which is close to what continuous networks could provide. Also, the equal amount of temporal-topological structure of patches is more efficient in future representation learning.

We also observe that the distribution of events tends to be uniform on the temporal patch with a smaller time interval as shown in Figure \ref{fig:interval}. The majority and minority of event numbers in the temporal patches with one-day intervals are 6087 and 3499 on Wikipedia, while the numbers are 28305 and 3651 of the one with five-day intervals. Ideally, if we split the raw representations by each moment, the distribution of events will be almost homogeneous, and no information loss. However, the computation will be much heavier in future representation learning due to an enormous number of patches. Thus, it is crucial to balance the number and amount of temporal-topological structure of patches, which is discussed in section \ref{ADE}.

\subsection{Multi-head Attention Mechanism}
The multi-head mechanism is also used to stabilize the learning process under the link prediction task. At the end of both the GSA and STT modules, the multi-head attention mechanism is adopted separately.

\textbf{The structural multi-head attention for GSA}
Since the multi-head attention mechanism is adopted at GSA, the final representation $h_u$ in section \ref{GSA} is concatenated with the output from each single-head in Equation \ref{multi1}.

\begin{equation}\label{multi1}
h_u = \  \bowtie ( h^{1}_{u}, h^{2}_{u}, \cdots h^{k}_{u})
\end{equation}

Where $\bowtie$ is the concatenate operation, and $k$ is the number of multiple heads. The graph structural attention heads share parameters across temporal patches.

\begin{figure}[h]
  \centering
  \includegraphics[width=\linewidth]{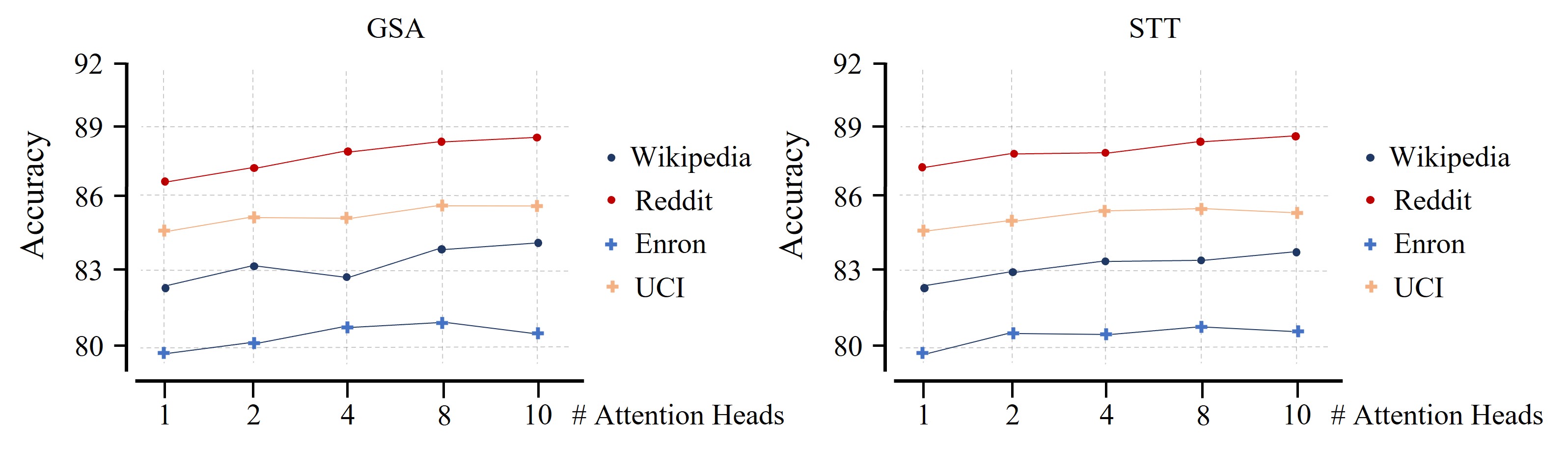}
  \caption{The accuracy of Efficient\_Dyn with different multi-heads on four datasets in the inductive link prediction tasks}
  \label{fig:multi}
\end{figure}

\textbf{The temporal multi-head attention for STT}
Similar with the above setting, the representation of each patch $z_{i}$ and relay $r(t)$ in section \ref{STT} are concatenated with the output from each single-head in Equation \ref{multi2} and \ref{multi3}.

\begin{equation}\label{multi2}
z_{i}(t) = \  \bowtie \left  ( z_{i}^{1}(t), z_{i}^{2}(t), \cdots z_{i}^{k}(t) \right)
\end{equation}

\begin{equation}\label{multi3}
r(t) = \  \bowtie \left  ( r^{1}(t), r^{2}(t), \cdots r^{k}(t) \right)
\end{equation}

In order to evaluate the contribution of the multi-head attention mechanism, we set a series of experiments for Efficient\_Dyn with different head numbers independently in the range {1, 2, 4, 8, 10} on two continuous and two discrete dynamic graph datasets. As shown in Figure \ref{fig:multi}, it can be observed that the accuracy of multi-head networks is better than the single-head networks on all four datasets. In addition, the accuracy does not keep increasing with a larger number of heads. The performance of the multi-head network stabilizes with eight attentions heads for both modules.

\subsection{Experimental Setup}
At the beginning of training, the Xavier \cite{glorot2010understanding} is to initialize the learnable parameters $W$ of each layer, which is to avoid the gradient from exploding or vanishing suddenly. The Gaussian Error Linear Unit (GELU) yields the final output non-linear representations at the end of GSA. The exponential linear unit (ELU) is utilized as the activation function for both temporal patches and relay in STT. A binary cross-entropy loss sends the probability distribution over predicted link prediction in both inductive and transductive tasks. The numbers of adaptive temporal patches of Reddit and Wikipedia are 16 and 12, finally. In addition, the dropout approach \cite{srivastava2014dropout} is introduced to avoid over-fitting during the training process, with the dropout rate in a range {$0.3$ to $0.7$}, which depends on the dataset and tasks. 

\bibliography{mybibfile}

\end{document}